\pdfoutput=1

\documentclass[11pt]{article}

\usepackage[final]{acl}

\usepackage{times}
\usepackage{latexsym}

\usepackage[T1]{fontenc}
\usepackage{CJKutf8}

\usepackage[utf8]{inputenc}

\usepackage{microtype}

\usepackage{inconsolata}

\usepackage{tabularx}
\usepackage{color}
\usepackage{amsthm}
\usepackage{amsmath}
\usepackage{graphicx,xcolor}
\theoremstyle{definition}

\title{Quantifying Memorization and Detecting Training Data of Pre-trained Language Models using Japanese Newspaper}

\author{Shotaro Ishihara \\
  Nikkei Inc.\\
  Tokyo, Japan \\
  \texttt{shotaro.ishihara@nex.nikkei.com} \\\And
  Hiromu Takahashi \\
  Independent Researcher \\
  Tokyo, Japan \\
  \texttt{hiromu.takahashi56@gmail.com} \\}

\begin{document}
\maketitle
\begin{abstract}
Dominant pre-trained language models (PLMs) have demonstrated the potential risk of memorizing and outputting the training data.
While this concern has been discussed mainly in English, it is also practically important to focus on domain-specific PLMs.
In this study, we pre-trained domain-specific GPT-2 models using a limited corpus of Japanese newspaper articles and evaluated their behavior.
Experiments replicated the empirical finding that memorization of PLMs is related to the duplication in the training data, model size, and prompt length, in Japanese the same as in previous English studies.
Furthermore, we attempted membership inference attacks, demonstrating that the training data can be detected even in Japanese, which is the same trend as in English.
The study warns that domain-specific PLMs, sometimes trained with valuable private data, can ``copy and paste'' on a large scale.\footnote{An early version of this study was accepted for 
non-archival track of the Fourth Workshop on Trustworthy Natural Language Processing~\citep{Ishihara2024-ib}.}
\end{abstract}

\begin{CJK}{UTF8}{ipxm}
\section{Introduction}
\label{sec:introduction}

As pre-trained language models (PLMs) have become increasingly practical, critical views on the memorization of PLMs are emerging in security and copyright \citep{Bender2021-ki, Bommasani2021-sc, Weidinger2022-nn}.
Prior research has indicated that neural networks have the property of unintentionally memorizing and outputting the training data~\citep{Carlini2019-pf, Carlini2021-xi, Carlini2022-wv, 10.1145/3543507.3583199, 10.5555/3618408.3620094}.
In particular, \citet{Carlini2021-xi} demonstrated that memorized personal information can be detected from GPT-2 models~\citep{radford2019language}.
This can lead to an invasion of privacy, reduced utility, and reduced ethical practices \citep{Carlini2022-wv}.
If there is no novelty in the generation, there would be a problem with copyright \citep{mccoy-etal-2023-much, Franceschelli2023-sx}.

Research on memorization of PLMs has been intensively advanced, and empirical findings have been reported \citep{ishihara-2023-training}.
Initial studies remain on the qualitative side~\citep{Carlini2021-xi}, and subsequent studies have begun to focus on quantitative evaluations.
According to one of the first comprehensive quantitative studies \citep{Carlini2022-wv}, the memorization of PLMs is strongly related to the string \emph{duplications} in the training set, \emph{model size}, and \emph{prompt length}.
Benchmarking of memorized string detection has also progressed, including constructing evaluation sets \cite{shi2024detecting, Duarte2024-bf, Kaneko2024-jx, Duan2024-zu}.

These studies were conducted in English, and their reproducibility is uncertain under domain-specific conditions.
Domain-specific PLMs are sometimes built on rare private corpora and have smaller pre-training corpora than general PLMs.
When the data size is small, models tend to be pre-trained in multiple epochs.
However, increasing the number of epochs is equivalent to string duplications, which risks increased memorization.
Furthermore, security and copyright considerations become increasingly important, as the memorized contents tend to be more specific than general corpora.
We, therefore, pose the following practically significant questions about domain-specific PLMs: \emph{how much of the pre-training data is memorized}, and \emph{is the memorized data detectable}?

This study is the first attempt to quantify the memorization of domain-specific PLMs using a limited corpus of Japanese newspaper articles.
Our research objective is \emph{to identify the memorization properties of domain-specific PLMs}.
First, we developed a framework for quantifying the memorization and detecting training data of PLMs using Japanese newspaper articles (Section~\ref{sec: proposed}).
We then pre-trained domain-specific GPT-2 models and quantified their memorization (Section~\ref{sec: experiments}).
Furthermore, we addressed membership inference attacks~\citep{Shokri2017-tr}, which predicts whether the output string was included in the training data (Section~\ref{sec: inference}).

The main findings and contributions of this paper are summarized as follows.
\begin{itemize}
    \item \textbf{Quantification}: Japanese PLMs were demonstrated to sometimes memorize and output the training data on a large scale. Experiments reported that memorization was related to duplication, model size, and prompt length. These empirical findings, which had been reported in English, were found for the first time in Japanese.
    \item \textbf{Detection}: Experiments demonstrated that the training data was detected from PLMs even in Japanese. The membership inference approach suggested in English was successful with the AUC (area under the ROC curve) score of approximately 0.6. As well as the empirical findings of memorization, the more duplicates and the longer the prompt, the easier the detection was.
\end{itemize}

\section{Related Work}
\label{sec: related}

This section reviews related work and highlights the position of this study.

\subsection{Memorization of PLMs}
\label{subsec: related_memorization}

Memorization of PLMs refers to the phenomenon of outputting fragments of the training data.
Research on memorization is diverse, with various definitions and assumptions.
We focus on autoregressive language models, such as the GPT family~\cite{Alec2018,radford2019language,Brown2020-cf,black-etal-2022-gpt}.
These are promising models and major research targets.

\paragraph{Definition of memorization.}

Many studies have adopted definitions based on partial matching of strings~\citep{Carlini2021-xi,Carlini2022-wv,pmlr-v162-kandpal22a}.
This definition of \emph{eidetic memorization} assumes that memorized data are extracted by providing appropriate prompts to PLMs.
Another definition of \emph{approximate memorization} considers string fuzziness.
For similarity, \citet{lee-etal-2022-deduplicating} used the token agreement rate, and \citet{ippolito-etal-2023-preventing} used BLEU.

Our study designed the first of these definitions in Japanese and reported the experimental results.
Both definitions of memorization are ambiguous in languages without obvious token delimiters such as Japanese.
Definitions based on the concepts of differential privacy~\citep{Jagielski2020-wl,Nasr2021-qu} and counterfactual memorization~\citep{Zhang2021-sh} are beyond the scope of this study.

\paragraph{Issues with memorization of PLMs.}

Training data extraction is a security attack related to the memorization of PLMs~\citep{ishihara-2023-training}.
Many studies follow the pioneering work of \citet{Carlini2021-xi}.
They reported that a large amount of information could be extracted by providing GPT-2 models with various prompts (generating candidates) and performing membership inference.
In particular, when dealing with PLMs with sensitive domain-specific information such as clinical data, the leakage of training data can lead to major problems~\citep{Nakamura2020-pg,lehman-etal-2021-bert, Jagannatha2021-pf, Singhal2022-dq, Yang2022-ct}.
It is also necessary to discuss from the perspective of human rights, such as the right to be forgotten~\citep{Li2018-ft, Ginart2019-jd, Garg2020-as}.

There has been a traditional research area for evaluating the quality of text generation, but few studies have focused on novelty \citep{mccoy-etal-2023-much}.
Novelty in text generation is directly related to the discussion of copyright~\citep{Franceschelli2023-sx}.
\citet{10.1145/3543507.3583199} analyzed plagiarism patterns in PLMs using English domain-specific corpora.

The memorization of PLMs has also been identified as data contamination damaging the integrity of the evaluation set.
Several studies have identified the inclusion of evaluation sets in the large datasets used for pre-training, which has led to unfairly high performance~\citep{magar-schwartz-2022-data, jacovi-etal-2023-stop, aiyappa-etal-2023-trust}.

Our study of quantifying memorization and performing membership inference would serve to confront these issues precisely in Japanese.

\subsection{Quantifying Memorization and Detecting Training Data of PLMs}
\label{subsec: related_paywall}

Recent studies have quantitatively evaluated memorization and related issues.

\paragraph{Empirical findings.}

As mentioned in Section~\ref{sec:introduction}, empirical findings in English are known that the memorization of PLMs is strongly related to the string duplications in the training set, model size, and prompt length \citep{Carlini2021-xi}.
There are supportive reports for this finding for duplication \citep{lee-etal-2022-deduplicating, Tirumala2022-qs, 10.1145/3543507.3583199, ippolito-etal-2023-preventing, pmlr-v162-kandpal22a, mccoy-etal-2023-much}, model size \cite{huang-etal-2022-large, pmlr-v162-kandpal22a, 10.1145/3543507.3583199, karamolegkou-etal-2023-copyright, ippolito-etal-2023-preventing, mccoy-etal-2023-much}, and prompt length \cite{huang-etal-2022-large, pmlr-v162-kandpal22a}.

\paragraph{Evaluation sets for quantification.}

\begin{figure}[ht]
  \centering
  \includegraphics[width=5cm]{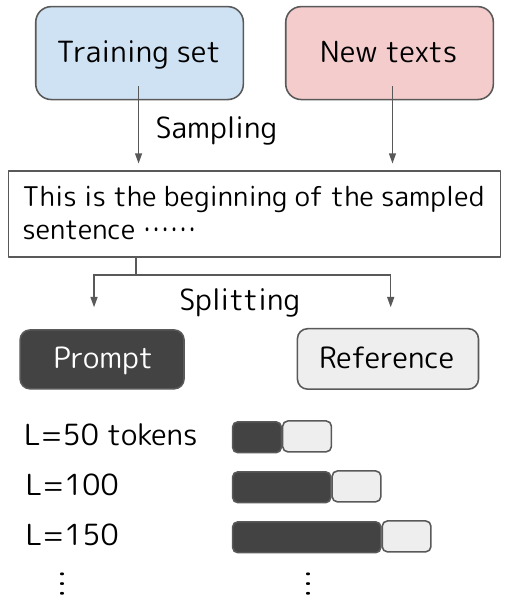}
  \caption{
    The existing method for constructing an evaluation set for quantifying memorization and detecting training data.
    This procedure requires sampling data from the training set used to pre-train and splitting the text into prompts and references.
    \textcolor{blue}{Positive} examples are created from training data and \textcolor{red}{negative} examples from new text that are guaranteed not to be training data.
  }
  \label{fig:evaluation_set}
\end{figure}

We describe the quantification methods used in the pioneering study~\citep{Carlini2022-wv} and point out the potential for improvement.
Owing to inference time limitations, it is impossible to evaluate memorization using all of the training data.
For example, \citet{Carlini2022-wv} targeted GPT-Neo models~\citep{black-etal-2022-gpt} and constructed an evaluation set by sampling 50,000 samples from the Pile dataset~\citep{Gao2020-kc} used for pre-training.
Sampling and string splitting are unavoidable during the construction of the evaluation set, as shown in Figure~\ref{fig:evaluation_set}.
Each sampled sentence was divided into prompts of each length from 50 to 500 tokens at the beginning, with the following 50 tokens as references.

\begin{figure*}[ht]
  \centering
  \includegraphics[width=14.2cm]{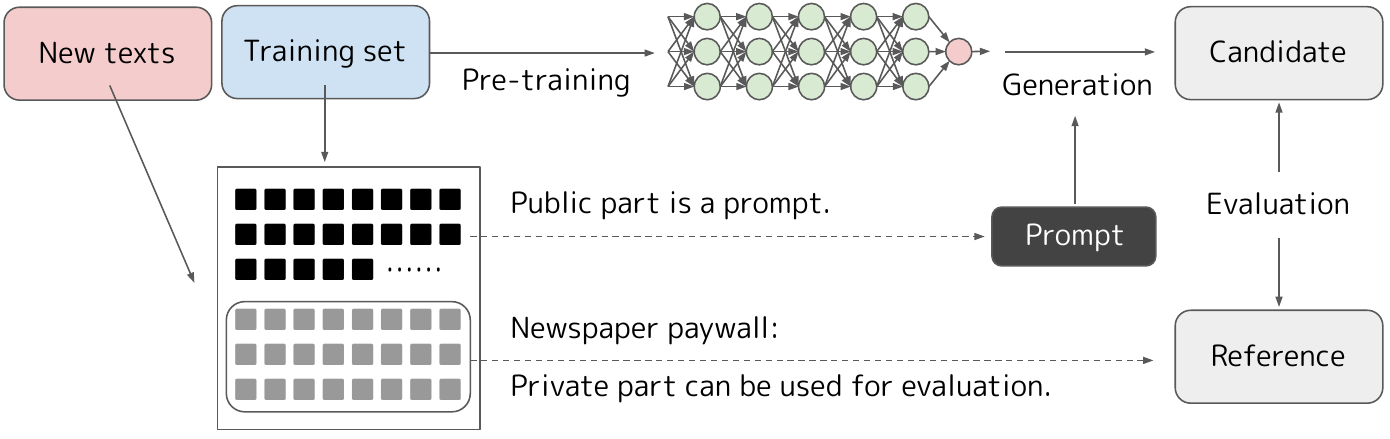}
  \caption{
    The procedure of quantifying the memorization and training data detection of PLMs in this study.
    First, we pre-trained GPT-2 models using newspaper articles as a training set.
    We then generated strings using the public part as a prompt.
    The memorization was quantified using the private part.
    We also tackle the training data detection task, using articles used for pre-training as \textcolor{blue}{positive} examples and not as \textcolor{red}{negative} examples.
  }
  \label{fig:overview}
\end{figure*}

However, this splitting does not consider the importance of references.
In other words, it does not consider whether references are protected subjects against security concerns.
We argue that using newspaper articles can provide real-world settings in data splitting via their paywalls.
Newspaper paywall restricts access to online content through a paid subscription \citep{myllylahti2016newspaper}.
Online news services with paid subscription plans often publish newspaper articles only at the beginning, with the rest of the text available only to their members.
This system creates a real-world setting in which there is a \emph{private part} following the \emph{public part} as illustrated in Figure~\ref{fig:overview}.
Using private parts as references can achieve the splitting in which publishers hide important information that they want to preserve.

Newspaper paywalls are often discussed in the literature tied to journalism.
For example, \citet{Kim2020-yc} examined the impact of newspaper paywalls on daily page views and differences among publishers.
Several other studies were conducted in the context of publishers' digital strategies \citep{Myllylahti2014-ks, Carson2015-gr, Sjovaag2016-bb}.

\paragraph{Evaluation sets for training data detection.}

To evaluate the detection of memorized training data from PLMs, it is necessary to have data that is guaranteed not to have been used for pre-training.
A promising approach is to use new texts generated after constructing PLMs.
\citet{shi2024detecting} constructed a dataset based on the creation date of the Wikipedia articles.
\citet{Duarte2024-bf} developed a dataset from the publication years of 165 books.

Along with evaluation sets, detection methods have been explored.
For example, \citet{shi2024detecting} proposed Min-$k\%$ Prob, which extracts $k$ \% tokens with high log-likelihood and uses the average log-likelihood for detection.
Min-$k\%$ Prob is regarded as one of the current prevailing methods~\citep{Kaneko2024-jx, Zhang2024-zo, meeus2024copyright}.
\citet{Kaneko2024-jx} introduced {SaMIA}, which generates multiple candidates and calculates the average of the ROUGE-1 \citep{lin-2004-rouge} without using the output of likelihood.
The AUC score and TPR@10\%FPR (True Positive Rate when False Positive Rate is 10 \%) are used as the metrics~\citep{mattern-etal-2023-membership, shi2024detecting, Kaneko2024-jx}.
Note that \citet{Carlini2021-xz} recommended reporting TPR when FPR is low in membership inference assessments.

We use Japanese newspaper articles to construct the evaluation set and perform the existing detection method.
Newspaper articles are generated daily, ensuring data is not used for pre-training.
Given the widespread use of newspaper articles in many languages, our proposal has the appeal of high versatility in low-resource languages.

\section{Problem Statement \& Methodology}
\label{sec: proposed}

This section explains the problem addressed in this study and the methodology (Figure~\ref{fig:overview}).
We use a methodology similar to that in \citet{Carlini2022-wv}.

\subsection{Constructing Evaluation sets.}

As described in Section~\ref{sec: related}, we construct evaluation sets using newspaper articles and paywalls.

\paragraph{Evaluation sets for quantification.}

To quantify memorization, sentences need to be split into prompts and references.
We propose to use the beginning of the newspaper article (the public part) as a prompt and the continuation in the paywall (the private part) as a reference.

\paragraph{Evaluation sets for training data detection.}

Positive and negative examples are required to measure the performance of training data detection.
We propose to use the newspaper articles used to construct the PLMs as positive examples and those published later as negative examples.

\subsection{Quantifying Memorization}
\label{subsec: statement_memorization}

The three steps to quantify memorization are described.

\paragraph{Step 1. Preparing PLMs.}

First, as a preparation, PLMs are built using all sentences containing both public and private parts of newspaper articles.

\paragraph{Step 2. Generating candidate.}

For a given article in the evaluation set, we consider the string in the public part to be prompt and generate a string that follows.

\paragraph{Step 3. Calculating similarity.}

The degree of memorization is evaluated by comparing the generated string with the private part.
We designed two Japanese definitions of memorization of PLMs.
While previous studies were based on English words, we must consider that there are no spaces between words in Japanese.
The definitions of memorization in this study are as follows.
\begin{itemize}
    \item The eidetic memorization is measured by the number of forward-matching characters. This is a definition that is independent of the properties of the word segmenter and tokenizer. Therefore, it has advantages in dealing with languages without explicit word boundaries, such as Japanese. As this study uses Japanese newspaper articles and their paywall, we had to use a derivation slightly different from the original eidetic memorization. It is a derivation of the original definition with the restriction of forward-matching characters.
    \item The approximate memorization is measured by a normalized Levenshtein distance~\citep{Yujian2007-qk}. The Levenshtein distance is a measure of the number of characters required to match one string to the other. We convert this value to similarity by dividing it by the number of characters of the higher value.
\end{itemize}

\subsection{Detecting Training Data.}
\label{subsec: statement_detecting}

We also attempt to detect memorized training data.
In this problem setting, there are two differences from quantifying memorization.
\begin{itemize}
    \item The reference is not available. This is because the situation where an attacker knows the reference is not realistic.
    \item The likelihood of PLMs is available. We can get not only the output string but also the likelihood.
\end{itemize}

Therefore, instead of Step 3 in which memorization is quantified in terms of string similarity between the candidate and the reference, we establish Step 3' in which membership probability is calculated.

\paragraph{Step 3'. Calculating membership probability.}

For the detection method, we use Min-$k\%$ Prob for $k$ in $\{10, 20, 30, 40, 50, 60\}$.
As described in Section~\ref{sec: related}, Min-$k\%$ Prob calculates the membership probability by extracting and averaging $k$ \% tokens with high log-likelihood.
The AUC score and TPR@10\%FPR are reported in common with the previous studies.

\begin{table*}[t]
\centering
\small
\begin{tabular}{lrrclc}
\hline
model name & parameter size & \multicolumn{2}{l}{eidetic} & \multicolumn{2}{l}{approximate} \\ \hline
aggregation & - & max & average & average & median \\ \hline
\texttt{gpt2-nikkei-1epoch} & 0.1 B & 25 & 0.560 & 0.190537 & 0.120345 \\
\texttt{gpt2-nikkei-5epoch} & 0.1 B & 25 & 0.839 & 0.229408 & 0.142857 \\
\texttt{gpt2-nikkei-15epoch} & 0.1 B & \textbf{48} & 0.788 & 0.236079 & 0.142857 \\
\texttt{gpt2-nikkei-30epoch} & 0.1 B & \textbf{48} & \textbf{0.948} & \textbf{0.241923} & \textbf{0.149627}\\
\texttt{gpt2-nikkei-60epoch} & 0.1 B & \textbf{48} & 0.874 & 0.238184 & 0.145833 \\ \hline
\texttt{rinna/japanese-gpt2-small} & 0.1 B & 12 & 0.580 & 0.181397 & 0.115385 \\
\texttt{rinna/japanese-gpt2-medium} & 0.3 B & 15 & 0.657 & 0.205017 & 0.129032 \\
\texttt{abeja/gpt2-large-japanese} & 0.7 B & 19 & 0.760 & 0.210954 & 0.136364 \\
\texttt{rinna/japanese-gpt-1b} & 1.3 B & 18 & 0.882 & 0.219001 & 0.142857 \\ \hline
\end{tabular}
\caption{
    Experimental results of memorization for each model.
    As the number of epochs increases, memorization enhances.
    The domain-specific GPT-2 models memorized their training data more than the other models.
    The memorization of general GPT-2 models increased along with the parameter size.
    The parameter size B stands for Billion.
}
\label{tab:experiment}
\end{table*}

\section{Experiment 1: Quantification}
\label{sec: experiments}

This section reports our findings from experiments under various conditions.
First, multiple PLMs and the evaluation set were prepared, and then memorization was quantified.
We analyzed the results from a quantitative and qualitative perspective.

\subsection{Preparing Evaluation Set}

As a dataset containing information on newspaper paywalls, we selected the corpus of Japanese newspaper articles provided by Nikkei Inc\footnote{\url{https://aws.amazon.com/marketplace/seller-profile?id=c8d5bf8a-8f54-4b64-af39-dbc4aca94384}}.
The newspaper articles were covered from March 23, 2010\footnote{Launch date of Nikkei's online edition} to December 31, 2021.
In this corpus, the shorter of the first 200 words or half the number of words in the entire article is defined as the public part.
This corpus was filtered to include approximately 1-2 billion (B) tokens.
Note that there are cases in which the entire article, including the private part, is made public according to various circumstances such as the importance of the topics.

\begin{figure}[ht]
  \centering
  \includegraphics[width=6cm]{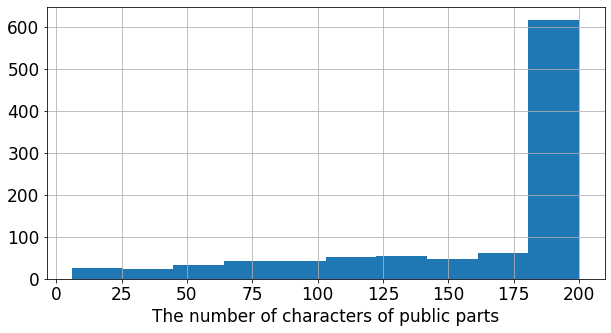}
  \caption{
    Histogram of the number of characters in the public part in the evaluation set.
    Most articles are around 200 words, but some are shorter.
  }
  \label{fig:public_stats}
\end{figure}

\begin{figure}[ht]
  \centering
  \includegraphics[width=6cm]{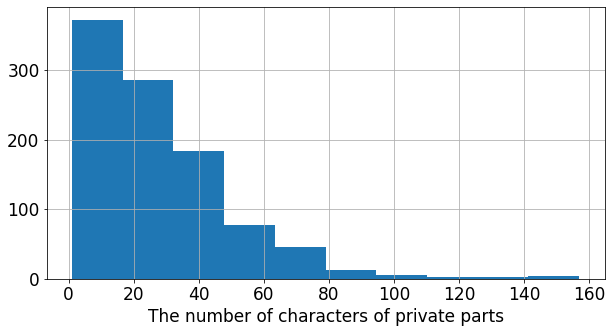}
  \caption{
    Histogram of the number of characters up to the end of the first sentence in the private part of the evaluation set.
    Nine articles exceeded 200 characters and were therefore skipped in the visualization.
  }
  \label{fig:private_stats}
\end{figure}

We randomly sampled 1,000 articles published in 2021 as our evaluation set.
The number of characters in the public part was approximately 200 words in most articles; however, some were shorter.
Only a minority (25 articles) ended the public part using punctuation marks\footnote{Japanese punctuation mark is ``。''.}.
The private parts are extremely long for some articles, and we extracted them until the end of the first sentence\footnote{We used \texttt{bunkai} (\url{https://github.com/megagonlabs/bunkai}).} to simplify the problem.
Histograms of the number of characters in the public and private part in the constructed evaluation set are shown in Figure~\ref{fig:public_stats} and \ref{fig:private_stats}.

\subsection{Step 1: Preparing PLMs}

For comparison, we used both domain-specific and general GPT-2 models in our experiments.

\paragraph{Domain-specific GPT-2.}

The domain-specific GPT-2 models were pre-trained using the full text of the corpus.
The parameter size is 0.1 B (117 million).
The model was saved for multiple training epochs: 1, 5, 15, 30, and 60.
In the pre-training of the domain-specific GPT-2 models, the loss to the validation set was 3.33 at 20 epochs, dropping to 3.30 at 40 epochs and slightly worse to 3.35 at 60 epochs.
We stopped the pre-training at 60 epochs due to this observed loss.
The articles in the evaluation set were also included in the corpus.
A list of models can be found in Table\ref{tab:experiment}, where \texttt{gpt2-nikkei-\{X\}epoch} is the model trained for \texttt{X} epochs.

Previous research in English~\cite{Carlini2022-wv} using models from 0.1 B to 6 B identified comparable trends in training data overlap and prompt length across all models.
Therefore, we consider the experiments with the 0.1 B worthwhile.
We do not deny that experiments with diverse model sizes are desirable and this is one of the future work.

We used Hugging Face Transformers~\citep{wolf-etal-2020-transformers} for pre-training\footnote{We used Transformers 4.11 and TensorFlow 2.5.} and the unigram language model~\citep{kudo-2018-subword} as the tokenizer.
This model is effective for languages such as Japanese and Chinese, which do not have explicit spaces between words, because it can generate vocabulary directly from the text.
The vocabulary size was 32,000.
The hyperparameters were set up with reference to the Transformers document\footnote{\url{https://github.com/huggingface/transformers/tree/main/examples/flax/language-modeling}}.
Specifically, we set the learning rate to 0.005, batch size to 64, weight decay~\citep{loshchilov2018decoupled} to 0.01, and the optimization algorithm to Adafactor~\citep{Shazeer2018-tu}.
Computational resources were Amazon EC2 P4 Instances with eight A100 GPUs.

\paragraph{General GPT-2.}

Models pre-trained on different datasets were also included for comparison.
This is because it is possible for the strings generated to coincide by chance, regardless of the nature of the memorization.
We selected models with parameter sizes of 0.1, 0.3, 0.7, and 1.3 B.
The model names in Table~\ref{tab:experiment} are the public names of the Hugging Face Models\footnote{\url{https://huggingface.co/models}}.
The models were pre-trained on the Japanese Wikipedia\footnote{\url{https://meta.wikimedia.org/wiki/Data_dumps}} and CC-100\footnote{\url{https://data.statmt.org/cc-100/}}.

\subsection{Step 2: Generating Candidate}

We generated a single string from a single prompt using a greedy method that produces the word with the highest conditional probability each time.
Exploring decoding strategies is one of the research questions for the future.

\subsection{Step 3: Calculating Similarity \& Quantitative Analysis}

\begin{table}[t]
\centering
\small
\begin{tabular}{rrr}
\hline
prompt length & eidetic & approximate \\ \hline
-116 & 0.892157	& 0.235276\\
116-187 & 1.010101 & 0.279301 \\
187-198 & 0.734694 & 0.224895\\
198-199 & 0.864865 & 0.216248\\
199-200 & \textbf{1.454545} & \textbf{0.295147}\\ \hline
\end{tabular}
\caption{
    Average eidetic and approximate memorization when the evaluation set was divided into 200 samples.
    The chunk with the longest prompts had the largest memorization for the model of 60 epochs.
}
\label{tab:context}
\end{table}

\begin{table*}[t]
\centering
\small
\begin{tabularx}{\linewidth}{|l|X|c|c|}
\hline
public / private / model name & strings & eidetic & approximate \\ \hline
public part & (...) 年明け以降の新型コロナウイルスの新規感染者数が大幅に増加するとの懸念が一定の重荷になっている。
[EN] (...) There is a certain burden of concern that the number of new cases of COVID-19 will increase significantly after the new year. & - & - \\ \hline
private part & 前引け後の東証の立会外で、国内外の大口投資家が複数の銘柄をまとめて売買する「バスケット取引」は約65億円成立した。 [EN] Approximately 6.5 billion yen in ``basket trading,'' in which large investors from Japan and abroad buy and sell multiple stocks at once, was concluded outside the TSE auction after the previous close. & - & - \\ \hline
\texttt{gpt2-nikkei-1epoch} & JPX日経インデックス400と東証株価指数(TOPIX)も下落している。 & 0 & 0.052632 \\ \hline
\texttt{gpt2-nikkei-5epoch} & 市場からは「きょうは2万9000円～2万9000円の範囲で、この水準を上抜けるには戻り待ちの売りが出やすい」(国内証券ストラテジスト)との声があった。 & 0 & 0.093333 \\ \hline
\texttt{gpt2-nikkei-15epoch} & \colorbox{green}{前引け後の東証の立会外で、国内外の大口投資家が複} \colorbox{green}{数の銘柄をまとめて売買する「バスケット取引」は約} 396億円成立した。 & 48 & 0.948276 \\ \hline
\texttt{gpt2-nikkei-30epoch} & \colorbox{green}{前引け後の東証の立会外で、国内外の大口投資家が複} \colorbox{green}{数の銘柄をまとめて売買する「バスケット取引」は約} 412億円成立した。 & 48 & 0.948276 \\ \hline
\texttt{gpt2-nikkei-60epoch} & \colorbox{green}{前引け後の東証の立会外で、国内外の大口投資家が複} \colorbox{green}{数の銘柄をまとめて売買する「バスケット取引」は約} 344億円成立した。 & 48 & 0.948276 \\ \hline
\texttt{rinna/japanese-gpt2-small} & 日経平均株価は前日比100円程度安の2万8800円近辺で軟調に推移している。 & 0 & 0.035088 \\ \hline
\texttt{rinna/japanese-gpt2-medium} & 日経平均株価は、前日比100円程度安の2万8800円近辺で軟調に推移している。 & 0 & 0.052632 \\ \hline
\texttt{abeja/gpt2-large-japanese} & 日経平均株価は、前日比100円程度安の2万8800円近辺で軟調に推移している。 & 0 & 0.052632 \\ \hline
\texttt{rinna/japanese-gpt-1b} & \texttt{</s>} & 0 & 0.000000 \\ \hline
\end{tabularx}
\caption{
    The sample in the evaluation set with the highest eidetic memorization in \texttt{gpt2-nikkei-60epoch} and the generated results.
    Strings that forward match the private part for reference are highlighted in \colorbox{green}{green}.
}
\label{tab:example}
\end{table*}

\begin{table*}[ht]
\centering
\small
\begin{tabular}{llrrrrrrrrrrr}
\hline
& & \multicolumn{5}{c}{AUC} & & \multicolumn{5}{c}{TPR@10\%FPR} \\
\cline{3-7} \cline{9-13}
method & model name & 32 & 64 & 128 & 256 & 512 & & 32 & 64 & 128 & 256 & 512 \\ \hline
Min-$k\%$ Prob & \texttt{gpt2-nikkei-1epoch} & 0.50 & 0.53 & 0.55 & 0.55 & 0.56 & & 18.5 & 21.7 & 21.9 & 20.1 & 19.6 \\
($k=10$) & \texttt{gpt2-nikkei-5epoch} & \textbf{0.51} & \textbf{0.55} & 0.59 & 0.58 & 0.58 & & 19.1 & \textbf{23.7} & 26.7 & 25.7 & 20.9 \\
& \texttt{gpt2-nikkei-15epoch} & 0.50 & 0.54 & 0.59 & 0.59 & 0.59 & & \textbf{19.6} & 22.5 & 26.9 & 24.8 & \textbf{23.4} \\
& \texttt{gpt2-nikkei-30epoch} & 0.50 & 0.53 & 0.58 & 0.59 & \textbf{0.60} & & 16.8 & 21.0 & 25.9 & \textbf{25.7} & 19.6 \\
& \texttt{gpt2-nikkei-60epoch} & 0.50 & 0.54 & \textbf{0.60} & \textbf{0.60} & 0.59 & & 15.8 & 21.0 & \textbf{27.6} & 25.0 & 19.6 \\ \hline
Min-$k\%$ Prob & \texttt{gpt2-nikkei-1epoch} & 0.46 & 0.47 & 0.48 & 0.50 & 0.53 & &  11.4 & 15.0 & 15.0 & 17.3 & 14.9 \\
($k=20$) & \texttt{gpt2-nikkei-5epoch} & 0.48 & 0.50 & 0.52 & 0.53 & 0.55 & & 13.7 & 19.5 & 18.1 & 18.8 & 17.4 \\
& \texttt{gpt2-nikkei-15epoch} & 0.46 & 0.49 & 0.53 & 0.54 & 0.56 & & 12.6 & 19.7 & 20.7 & 20.6 & 18.3 \\
& \texttt{gpt2-nikkei-30epoch} & 0.45 & 0.48 & 0.52 & 0.54 & 0.58 & & 11.7 & 18.7 & 20.2 & 20.1 & 14.5  \\
& \texttt{gpt2-nikkei-60epoch} & 0.47 & 0.50 & 0.56 & 0.57 & 0.57 & & 13.1 & 18.9 & 23.8 & 23.0 & 17.9 \\ \hline
\end{tabular}
\caption{
    The performance (AUC and TPR@10\%FPR) of Min-$k\%$ Prob for $k=10$ and $k=20$ with the prompt length in $\{32, 64, 128, 256, 512\}$.
    Bold text means the best value in each column.
}
\label{tab:detection}
\end{table*}

\begin{figure}[ht]
  \centering
  \includegraphics[width=7.4cm]{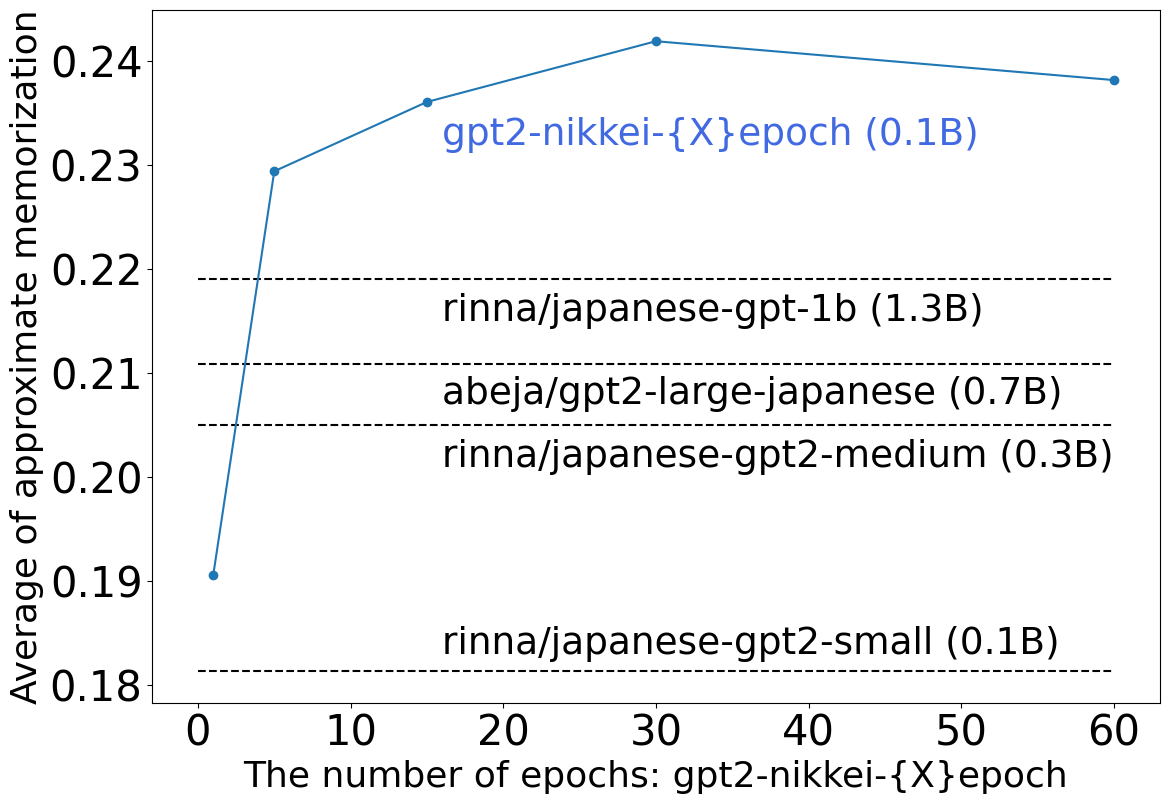}
  \caption{
    Visualization of the average value of approximate memorization. Similar results were confirmed for other metrics.
  }
  \label{fig:experimental_result}
\end{figure}

For all models, we computed the eidetic and approximate memorization of 1,000 articles in the evaluation set (Table~\ref{tab:experiment}).
For clarity, we illustrate the change in approximate memorization with each epoch in the domain-specific GPT-2 models in Figure~\ref{fig:experimental_result}.
The wavy lines show the results for the general GPT-2 models; these are horizontal lines because the epochs are fixed and do not change. 

Although the model at 30 epochs can not be regarded as overfitted, a large memorization was observed.
A previous study~\cite{Tirumala2022-qs} also reported the memorization of PLMs could occur before the overfitting.
The low average value is due to the large number of samples where no memorization is observed.

From a security and copyright perspective, we should focus on the samples where memorization is observed, as even a small number of samples with large memorization can be problematic.
Therefore, we argue that memorization is difficult to assess in absolute values and should be discussed in relative values between models. 

\paragraph{Memorization enhances along with epochs.}

This phenomenon replicates the empirical finding that memorization is associated with duplication within a training set, even in Japanese.
Figure~\ref{fig:experimental_result} shows that the median approximate memorization was strengthened through repeated pre-training on the same dataset.
As shown in Table~\ref{tab:experiment}, similar results were obtained for other metrics.
The maximum eidetic memorization changed from 25 to 48 after 15 epochs.
The average eidetic and approximate memorization also tended to increase in the epochs.
We speculate that the reason for the decreased memorization at the end of the epochs is due to the size of the model and training set.
Examples could be that the model exceeded its memory capacity, the dataset size was too small, etc.

\paragraph{The larger the size, the more memorized.}

In the other models, a larger number of parameters led to increased memorization.
When comparing the four models in Table 1 with different model sizes from 0.1 to 1.3 B, all metrics demonstrated an increase with size.
We speculate that this is because the general memorization property increases with an increasing number of parameters.
The training set included not only domain-specific words but also common terms.

\paragraph{The longer the context, the more memorized.}

To examine the effect of the length of the public part on memorization, we divided the evaluation set into 200 samples (Table~\ref{tab:context}).
Many samples were close to 200 in length, with thresholds of 116, 187, 198, and 199 in decreasing order.
The chunks with more characters had the largest average for both eidetic and approximate memorization for the model of 60 epochs.
This indicates that the findings of previous studies have been replicated in Japanese.

\paragraph{Domain-specific models do memorize.}

The domain-specific GPT-2 model recorded eidetic memorization of up to 25 characters in only one epoch.
This was higher than those of the other models at 0.3, 0.7, and 1.3 B.
The average eidetic and approximate memorization also exceeded those of the other models.
This indicates the training data were memorized, rather than a simple coincidence.

\subsection{Qualitative Analysis}

As a qualitative analysis, we report on a sample with the longest strings memorized in the evaluation set (Table~\ref{tab:example}).
In the generated results for each model, the strings that forward match the private part for reference are highlighted in \colorbox{green}{green}.
The full text can be found in the footnote URL
\footnote{\url{https://www.nikkei.com/article/DGXZASS0ISS14_Q1A231C2000000}}.

48 characters were memorized in the domain-specific GPT-2 model of 15 epochs.
This memorization persisted after 30 or 60 epochs.
The memorized pattern appeared only once in the training set.
The sudden loss drop in a particular sample is a phenomenon of memorization of PLMs, which has also been reported in \citet{Carlini2021-xi}.
No such phenomena were observed in the other models.
\texttt{rinna/japanese-gpt-1b} output a special token \texttt{</s>} indicating the end of a sentence, possibly due to a punctuation mark at the end of the public part.
Appendix \ref{sec:secondsample} shows a sample of the second-longest memorization, presenting an example where the public part does not end with punctuation.

\section{Experiment 2: Detection}
\label{sec: inference}

This section demonstrates that memorized strings can be detected from Japanese PLMs.
Specifically, we investigated whether detecting training data from Japanese PLMs is possible using the proven Min-$k$\% Prob in English.
We targeted the domain-specific GPT-2 models (1, 5, 15, 30, and 60 epochs) described in the previous section.

\subsection{Preparation Evaluation Set}

As explained in Section~\ref{subsec: statement_detecting}, newspaper articles published after pre-training were prepared as negative examples.
Specifically, we extracted 1,000 articles published in January 2023.
In summary, the evaluation set contained 1,000 articles in the pre-training data (used in the previous section) and 1,000 articles that were not used.
Each article was split into prompts and references with the prompt length in $\{32, 64, 128, 256, 512\}$, according to \citet{shi2024detecting}\footnote{Previous studies had not covered prompt lengths of 512, but we tried. This was because the newspaper articles had relatively long sentences.}.
The texts were split into words following the previous studies \citep{shi2024detecting, Kaneko2024-jx}.
We used MeCab~\cite{Kudo2006-lj} and mecab-ipadic-NEologd \cite{sato2017mecabipadicneologdnlp2017}.
Note that languages without explicit word-separation spaces, such as Japanese, require specific libraries and dictionaries.
The final number of positive and negative examples, truncated for data of insufficient length, was as follows: (957, 931) at 32-word counts, (908, 868) at 64, (772, 701) at 128, (452, 435) at 256, and (235, 237) at 512.

\subsection{Step 3': Calculating Membership Probability \& Quantitative Analysis}

Quantitative results demonstrated that training data is detectable in PLMs, even in Japanese.
The performance (AUC and TPR@10\%FPR) of Min-$k\%$ Prob for $k=10$ and $k=20$ with the prompt length in $\{32, 64, 128, 256, 512\}$ is shown in Table~\ref{tab:detection}.
We focus on $k=10$ from our search, which gave the best results (Appendix~\ref{sec:minkprob}).
The AUC scores exceeded the value of the random prediction (0.50) in almost all cases.
On the other hand, the $k=20$, which \citet{shi2024detecting} reported as the best, did not show sufficient performance.
This suggests the importance of the parameter $k$.
In summary, detection performance was related to duplication and prompt length, which is consistent with empirical findings on memorization.
As all model sizes are the same, their effects were outside the scope.

\paragraph{The more epochs, the more detectable.}

As the number of epochs increased, detection performance also improved.
In particular, values were larger in all columns when comparing epochs 1 and 5.

\paragraph{The longer the context, the more detectable.}

The AUC score and TPR@10\%FPR tended to increase as the prompt length was increased.
The prompt length of 32 had almost no detection performance, but when the prompt length reached 128, the AUC score approached 0.60.
It is worth highlighting that this AUC score was not high enough.
\citet{meeus2024copyright} pointed out that detection by Min-$k\%$ Prob does not work if the model size and the corpus size are not large.

\section{Conclusion}
\label{sec: discussion}

This study is the first attempt to quantify the memorization and detect training data of domain-specific PLMs that are not English but Japanese.
Although our study has some limitations, this is a major step forward, as there is even a scant discussion of string similarity concerning the memorization of domain-specific PLMs.

\subsection{Limitations}
\label{subsec:limitaitons}

Our study has some limitations.

\paragraph{Dataset accessibility.}

This study used newspaper articles with paywall characteristics.
The dataset is available for purchase, but not everyone has free access to it.
While this counterpart has the advantage of dealing with data contamination, there are disadvantages in terms of research reproducibility.

\paragraph{Larger evaluation sets and models.}

Although we randomly selected 1,000 articles as the evaluation set, experiments with a larger dataset are one of the prospects.
Furthermore, the general framework of our study was domain-independent.
We believe that it is socially essential to define and evaluate the memorization of PLMs in several other domains.
There is the potential for larger model sizes.
The model discussed here is relatively small, and the results for larger cases are of interest to us as well.

\paragraph{Association with danger.}

The security and copyright arguments are certainly not fully tested in the experiments of this study.
Considering the degree of danger of memorized strings is also important.
For example, the undesirable memorization of personally identifiable information (PII) such as telephone numbers and email addresses must be separated from acceptable memorization.
Several studies have evaluated the ability of PLMs to associate memorization with PII~\citep{huang-etal-2022-large, Shao2023-zk}.

\paragraph{Decoding strategy.}

In this study, a single string was generated from a single prompt using the greedy method, whereas the previous study~\citep{Carlini2021-xi, pmlr-v162-kandpal22a, lee-etal-2022-deduplicating} used various decoding strategies, such as top-k sampling, and tuned the temperature to increase the diversity of the generated texts.
\citet{Carlini2022-wv} reported that the choice of the decoding strategy does not considerably affect their experimental results.
By contrast, \citet{10.1145/3543507.3583199} observed that top-k and top-p sampling tended to extract more training data.

\paragraph{Measures for memorization.}

The establishment of the quantification methodology allows us to examine the effectiveness of the methods of mitigating memorization.
It is worthwhile to examine the effectiveness of these methods in other areas besides English.
\citet{ishihara-2023-training} classified defensive approaches into three phases:
\begin{itemize}
  \setlength{\parskip}{0cm}
  \item pre-processing: data sanitization~\citep{Ren2016-rt,Continella2017-gz,vakili-etal-2022-downstream}, and data deduplication~\citep{Allamanis2019-mr,pmlr-v162-kandpal22a,lee-etal-2022-deduplicating}.
  \item training: differential privacy~\citep{Yu2021-sb,Yu2021-uq,Li2021-ny,he2023exploring}, and information bottleneck~\citep{alemi2017deep,henderson2023a}.
  \item post-processing: confidence masking, and filtering\citep{perez-etal-2022-red}.
\end{itemize}

\section*{Ethics Statement}

This study involves training data extraction from PLMs, which is a security attack.
However, it is of course not intended to encourage these attacks.
Rather, we propose a framework for sound discussion to mitigate the dangers.
Although our study focused on Japanese, the findings can be easily applied to other languages.
This advantage is important for encouraging the development of PLMs worldwide.

The dataset used in this study was provided through appropriate channels by Nikkei Inc.
We have not engaged in any ethical or rights-issue data acquisition, such as scraping behind a paywall.
Many publishers provide article data for academic purposes, subject to payment of money and compliance with the intended use.
Therefore, we believe that our proposal is reproducible.

We used one AWS p4d.24xlarge instance\footnote{\url{https://aws.amazon.com/ec2/instance-types/p4/}} for 45 hours to pre-train the GPT-2 model for 60 epochs.

\paragraph*{\textit{Supplementary Materials Availability Statement:}}

We declare the Resource Availability in this paper as follows:

\begin{itemize}
    \item The corpus of Japanese newspaper articles was provided by Nikkei Inc\footnote{\url{https://aws.amazon.com/marketplace/seller-profile?id=c8d5bf8a-8f54-4b64-af39-dbc4aca94384}}.
    \item Source code of pre-training GPT-2 models\footnote{\url{https://github.com/huggingface/transformers/tree/main/examples/flax/language-modeling}} and Min-$k\%$ Prob\footnote{\url{https://github.com/swj0419/detect-pretrain-code}} is available from GitHub.
\end{itemize}

\section*{Acknowledgements}
We thank anonymous reviewers in INLG 2024 for their insightful comments and suggestions.
In addition, we express our gratitude to those involved in the review and discussions of the earlier versions of this study.


\appendix

\section{Sample of The Second Longest Memorization}
\label{sec:secondsample}

\begin{table*}[ht]
\centering
\small
\begin{tabularx}{\linewidth}{|l|X|c|c|}
\hline
public / private / model name & strings & eidetic & approximate \\ \hline
public part & (...) 日本政府は4月、30年度に温暖化ガス排出を13年度比46％減らす目標を打ち出した。秋に開かれた第26 [EN] (...) In April, the Japanese government set a target to reduce greenhouse gas emissions by 46 \% in FY30 compared to FY13. The 26th & - & - \\ \hline
private part & 回国連気候変動枠組み条約締約国会議（COP26）では、「世界の平均気温の上昇を1.5度に抑える努力を追求することを決意する」ことで合意した。 [EN] Conference of the Parties to the United Nations Framework Convention on Climate Change (COP26) agreed to ``resolve to pursue efforts to limit the increase in global average temperature to 1.5 degrees Celsius.'' & - & - \\ \hline
\texttt{gpt2-nikkei-1epoch} & \colorbox{green}{回国連気候変動枠組み条約締約国会議(COP26)で}、脱炭素に向けた投資や脱炭素の戦略を練り直す。 & 25 & 0.414286 \\ \hline
\texttt{gpt2-nikkei-5epoch} & \colorbox{green}{回国連気候変動枠組み条約締約国会議(COP26)で}も、企業の対応が注目されそうだ。 & 25 & 0.400000 \\ \hline
\texttt{gpt2-nikkei-15epoch} & \colorbox{green}{回国連気候変動枠組み条約締約国会議(COP26)では、} 50年の実質ゼロに向けた道筋を議論。 & 27 & 0.442857 \\ \hline
\texttt{gpt2-nikkei-30epoch} & \colorbox{green}{回国連気候変動枠組み条約締約国会議(COP26)では、} 30年目標の前倒しが議論された。 & 27 & 0.428571 \\ \hline
\texttt{gpt2-nikkei-60epoch} & \colorbox{green}{回国連気候変動枠組み条約締約国会議(COP26)では、} 各国が脱炭素に向けた行動計画を策定する。 & 27 & 0.457143 \\ \hline
\texttt{rinna/japanese-gpt2-small} & \colorbox{green}{回}気候変動枠組条約締約国会議(cop24)では、cop24で排出削減目標が達成された企業を「排出削減企業」として認定した。 & 1 & 0.357143 \\ \hline
\texttt{rinna/japanese-gpt2-medium} & \colorbox{green}{回}気候変動枠組条約締約国会議(cop24)で、cop21の目標達成に向けた具体的な行動計画の策定が合意された。 & 1 & 0.342857 \\ \hline
\texttt{abeja/gpt2-large-japanese} & \colorbox{green}{回}先進国首脳会議(伊勢志摩サミット)で、日本は「2030年目標」を公表した。 & 1 & 0.114286 \\ \hline
\texttt{rinna/japanese-gpt-1b} & \colorbox{green}{回}気候変動枠組条約締約国会議(COP26)では、パリ協定の実施指針となる「パリ協定実施指針」が採択された。 & 1 & 0.414286 \\ \hline
\end{tabularx}
\caption{
    The sample in the evaluation set with the second highest eidetic memorization in \texttt{gpt2-nikkei-60epoch} and the generated results.
    Strings that forward match the private part for reference are highlighted in \colorbox{green}{green}.
}
\label{tab:additional-example1}
\end{table*}

Table~\ref{tab:additional-example1} presents an example where the public part does not end with punctuation.
The full text can be found in the footnote URL
\footnote{\url{https://www.nikkei.com/article/DGKKZO78866030Y1A221C2DTA000}}.
The general trend was the same: the eidetic and approximate memorization increased with the number of epochs, and the other models showed smaller memorization.
The string ``回国連気候変動枠組み条約締約国会議(COP26)'' following ``第26'' was generated by only one epoch pre-training.
This suggests that they remember how the event\footnote{The 26th session of the Conference of the Parties to the United Nations Framework Convention on Climate Change (COP 26)} was notated in a domain-specific corpus.

There were few grammatical errors in the generated results; however, there were some factually incorrect statements, in smaller-sized models.
For example, \texttt{rinna/japanese-gpt2-small} and \texttt{rinna/japanese-gpt2-medium} in Table~\ref{tab:additional-example1} included the abbreviation of cop24 and cop21.
This is an incorrect generation in a situation where the public part gives the context of ``第26'', which means ``26th'' in English.
\texttt{abeja/gpt2-large-japanese} generated a different event name than the private part.

\section{Results of Detecting Training Data}
\label{sec:minkprob}

Figure~\ref{tab:detection_all} shows the performance of Min-$k\%$ Prob for $k$ in $\{10, 20, 30, 40, 50, 60\}$ with the prompt length in $\{32, 64, 128, 256, 512\}$.
The bold text, meaning the best value in each column, was concentrated at $k=10$.
Therefore, results for $k=10$ were reported in Section~\ref{sec: inference}.
The same pattern was observed in the other $k$ results, where the AUC scores tended to correlate with prompt length and number of epochs.

\begin{table*}[ht]
\centering
\small
\begin{tabular}{llrrrrrrrrrrr}
\hline
& & \multicolumn{5}{c}{AUC} & & \multicolumn{5}{c}{TPR@10\%FPR} \\
\cline{3-7} \cline{9-13}
method & model name & 32 & 64 & 128 & 256 & 512 & & 32 & 64 & 128 & 256 & 512 \\ \hline
Min-$k\%$ Prob & \texttt{gpt2-nikkei-1epoch} & 0.50 & 0.53 & 0.55 & 0.55 & 0.56 & &  18.5 & 21.7 & 21.9 & 20.1 & 19.6 \\
($k=10$) & \texttt{gpt2-nikkei-5epoch} & \textbf{0.51} & \textbf{0.55} & 0.59 & 0.58 & 0.58 & & 19.1 & \textbf{23.7} & 26.7 & 25.7 & 20.9 \\
& \texttt{gpt2-nikkei-15epoch} & 0.50 & 0.54 & 0.59 & 0.59 & 0.59 & & \textbf{19.6} & 22.5 & 26.9 & 24.8 & \textbf{23.4} \\
& \texttt{gpt2-nikkei-30epoch} & 0.50 & 0.53 & 0.58 & 0.59 & \textbf{0.60} & & 16.8 & 21.0 & 25.9 & \textbf{25.7} & 19.6 \\
& \texttt{gpt2-nikkei-60epoch} & 0.50 & 0.54 & \textbf{0.60} & \textbf{0.60} & 0.59 & & 15.8 & 21.0 & \textbf{27.6} & 25.0 & 19.6 \\ \hline
Min-$k\%$ Prob & \texttt{gpt2-nikkei-1epoch} & 0.46 & 0.47 & 0.48 & 0.50 & 0.53 & &  11.4 & 15.0 & 15.0 & 17.3 & 14.9 \\
($k=20$) & \texttt{gpt2-nikkei-5epoch} & 0.48 & 0.50 & 0.52 & 0.53 & 0.55 & & 13.7 & 19.5 & 18.1 & 18.8 & 17.4 \\
& \texttt{gpt2-nikkei-15epoch} & 0.46 & 0.49 & 0.53 & 0.54 & 0.56 & & 12.6 & 19.7 & 20.7 & 20.6 & 18.3 \\
& \texttt{gpt2-nikkei-30epoch} & 0.45 & 0.48 & 0.52 & 0.54 & 0.58 & & 11.7 & 18.7 & 20.2 & 20.1 & 14.5  \\
& \texttt{gpt2-nikkei-60epoch} & 0.47 & 0.50 & 0.56 & 0.57 & 0.57 & & 13.1 & 18.9 & 23.8 & 23.0 & 17.9 \\ \hline
Min-$k\%$ Prob & \texttt{gpt2-nikkei-1epoch} & 0.43 & 0.44 & 0.45 & 0.48 & 0.52 & & 9.4 & 12.1 & 11.3 & 14.6 & 14.5 \\
($k=30$) & \texttt{gpt2-nikkei-5epoch} & 0.46 & 0.47 & 0.48 & 0.50 & 0.54 & & 11.1 & 14.6 & 13.1 & 16.2 & 15.3 \\
& \texttt{gpt2-nikkei-15epoch} & 0.44 & 0.47 & 0.49 & 0.51 & 0.55 & & 10.4 & 17.4 & 16.2 & 15.7 & 15.3 \\
& \texttt{gpt2-nikkei-30epoch} & 0.43 & 0.46 & 0.49 & 0.52 & 0.56 & & 10.9 & 16.2 & 14.9 & 15.5 & 15.7 \\
& \texttt{gpt2-nikkei-60epoch} & 0.45 & 0.48 & 0.53 & 0.54 & 0.56 & & 10.4 & 17.2 & 19.9 & 21.5 & 16.2 \\ \hline
Min-$k\%$ Prob & \texttt{gpt2-nikkei-1epoch} & 0.41 & 0.42 & 0.43 & 0.47 & 0.51 & & 8.9 & 11.2 & 8.7 & 13.9 & 12.3 \\
($k=40$) & \texttt{gpt2-nikkei-5epoch} & 0.44 & 0.45 & 0.46 & 0.48 & 0.53 & & 9.3 & 14.1 & 12.3 & 14.4 & 16.6 \\
& \texttt{gpt2-nikkei-15epoch} & 0.43 & 0.46 & 0.47 & 0.49 & 0.54 & & 9.0 & 14.8 & 12.6 & 15.3 & 13.6 \\
& \texttt{gpt2-nikkei-30epoch} & 0.42 & 0.45 & 0.47 & 0.50 & 0.55 & & 9.0 & 13.5 & 12.6 & 12.8 & 15.3 \\
& \texttt{gpt2-nikkei-60epoch} & 0.43 & 0.47 & 0.51 & 0.52 & 0.55 & & 9.8 & 16.3 & 17.6 & 18.1 & 16.6 \\ \hline
Min-$k\%$ Prob & \texttt{gpt2-nikkei-1epoch} & 0.40 & 0.41 & 0.41 & 0.46 & 0.51 & & 8.4 & 9.6 & 8.0 & 13.1 & 11.9 \\
($k=50$) & \texttt{gpt2-nikkei-5epoch}  & 0.43 & 0.44 & 0.44 & 0.47 & 0.52 & & 9.1 & 11.8 & 11.4 & 13.9 & 16.6 \\
& \texttt{gpt2-nikkei-15epoch} & 0.42 & 0.45 & 0.46 & 0.48 & 0.53 & & 9.9 & 12.8 & 12.0 & 13.7 & 14.5 \\
& \texttt{gpt2-nikkei-30epoch} & 0.41 & 0.44 & 0.45 & 0.48 & 0.54 & & 9.0 & 12.6 & 11.5 & 12.6 & 15.7 \\
& \texttt{gpt2-nikkei-60epoch} & 0.42 & 0.46 & 0.49 & 0.50 & 0.54 & & 10.1 & 16.3 & 16.2 & 16.8 & 14.9 \\ \hline
Min-$k\%$ Prob & \texttt{gpt2-nikkei-1epoch} & 0.40 & 0.40 & 0.40 & 0.46 & 0.51 & & 8.5 & 8.6 & 7.4 & 11.5 & 14.0 \\
($k=60$) & \texttt{gpt2-nikkei-5epoch} & 0.42 & 0.43 & 0.43 & 0.47 & 0.51 & & 9.0 & 11.1 & 10.5 & 12.4 & 16.2 \\
& \texttt{gpt2-nikkei-15epoch} & 0.41 & 0.44 & 0.45 & 0.47 & 0.52 & & 9.0 & 14.0 & 11.5 & 15.0 & 16.2 \\
& \texttt{gpt2-nikkei-30epoch} & 0.40 & 0.43 & 0.44 & 0.48 & 0.54 & & 8.9 & 11.1 & 11.0 & 13.5 & 15.7 \\
& \texttt{gpt2-nikkei-60epoch} & 0.41 & 0.45 & 0.48 & 0.49 & 0.53 & & 9.7 & 15.2 & 14.8 & 15.5 & 15.3 \\ \hline
\end{tabular}
\caption{
    The performance (AUC and TPR@10\%FPR) of Min-$k\%$ Prob for $k$ in $\{10, 20, 30, 40, 50, 60\}$ with the prompt length in $\{32, 64, 128, 256, 512\}$.
    Bold text means the best value in each column.
}
\label{tab:detection_all}
\end{table*}

\end{CJK}
\end{document}